\documentclass[10pt,twocolumn,letterpaper]{article}

\usepackage{cvpr}
\usepackage{times}
\usepackage{epsfig}
\usepackage{graphicx}
\usepackage{amsmath}
\usepackage{amssymb}
\usepackage{algorithm}
\usepackage[noend]{algpseudocode}
\usepackage{caption}
\usepackage{array,booktabs}
\usepackage{siunitx}
\usepackage[dvipsnames]{xcolor}

\usepackage[pagebackref=true,breaklinks=true,letterpaper=true,colorlinks,citecolor=blue,linkcolor=blue,bookmarks=false]{hyperref}

\cvprfinalcopy 

\def\cvprPaperID{1701} 

\ifcvprfinal\fi

\begin{document}

\title{Weakly-Supervised Semantic Segmentation via Sub-category Exploration}



\author{
		Yu-Ting Chang$^{1}$
		\hspace{0.15in} Qiaosong Wang$^2$
		\hspace{0.15in} Wei-Chih Hung$^1$ 
		\hspace{0.15in} Robinson  Piramuthu$^2$\\
		\hspace{0.15in} Yi-Hsuan Tsai$^3$
		\hspace{0.15in} Ming-Hsuan Yang$^{1,4}$
		\vspace{1mm}\\
		$^1$UC Merced \hspace{0.15in} $^2$eBay Inc.\hspace{0.15in} $^3$NEC Labs America \hspace{0.15in} $^4$Google Research
	}

\maketitle

\begin{abstract}
Existing weakly-supervised semantic segmentation methods using image-level annotations typically rely on initial responses to locate object regions.
However, such response maps generated by the classification network usually focus on discriminative object parts, due to the fact that the network does not need the entire object for optimizing the objective function.
To enforce the network to pay attention to other parts of an object, we propose a simple yet effective approach that introduces a self-supervised task by exploiting the sub-category information.
Specifically, we perform clustering on image features to generate pseudo sub-categories labels within each annotated parent class, and construct a sub-category objective to assign the network to a more challenging task.
By iteratively clustering image features, the training process does not limit itself to the most discriminative object parts, hence improving the quality of the response maps.
We conduct extensive analysis to validate the proposed method and show that our approach performs favorably against the state-of-the-art approaches.

\end{abstract}

\section{Introduction}
The goal of semantic segmentation is to assign a semantic category to each pixel in the image. It has been one of the most important tasks in computer vision that enjoys a wide range of applications such as image editing and scene understanding. 
Recently, deep convolutional neural network (CNN) based methods \cite{fcn_pami,deeplab,dilated} have been developed for semantic segmentation and achieved significant progress.
However, such approaches rely on learning supervised models that require pixel-wise annotations, which take extensive effort and time.  
To reduce the effort in annotating pixel-wise ground truth labels, numerous weakly-supervised methods are proposed using various types of labels such as image-level \cite{ahn2018learning,kolesnikov2016seed,pathak2015constrained,pinheiro2015image}, video-level \cite{Chen_IJCV_2020,Zhong_ACCV_2016,Tsai_ECCV_2016}, bounding box \cite{papandreou2015weakly,dai2015boxsup,khoreva_CVPR17}, point-level \cite{Bearman_ECCV16}, and scribble-based \cite{lin2016scribblesup,Vernaza_CVPR17} labels.
In this work, we focus on using image-level labels which can be obtained effortlessly, yet a more challenging case under the weakly-supervised setting.

Existing algorithms mainly consist of three sequential steps to perform weakly-supervised training on the image-level label: 1) predict an initial category-wise response map to localize the object, 2) refine the initial response as the pseudo ground truth, and 3) train the segmentation network based on pseudo labels.
Although promising results have been achieved by recent methods \cite{ahn2018learning,huang2018weakly,wang2018weakly,wei2018revisiting}, most of them focus on improving the second and the third steps. Therefore, these approaches may suffer from inaccurate predictions generated in the first step, i.e., initial response.
Here, we aim to improve the performance of initial predictions which will benefit succeeding steps.

\begin{figure}[t]
	\centering
	\includegraphics[width=1\linewidth]{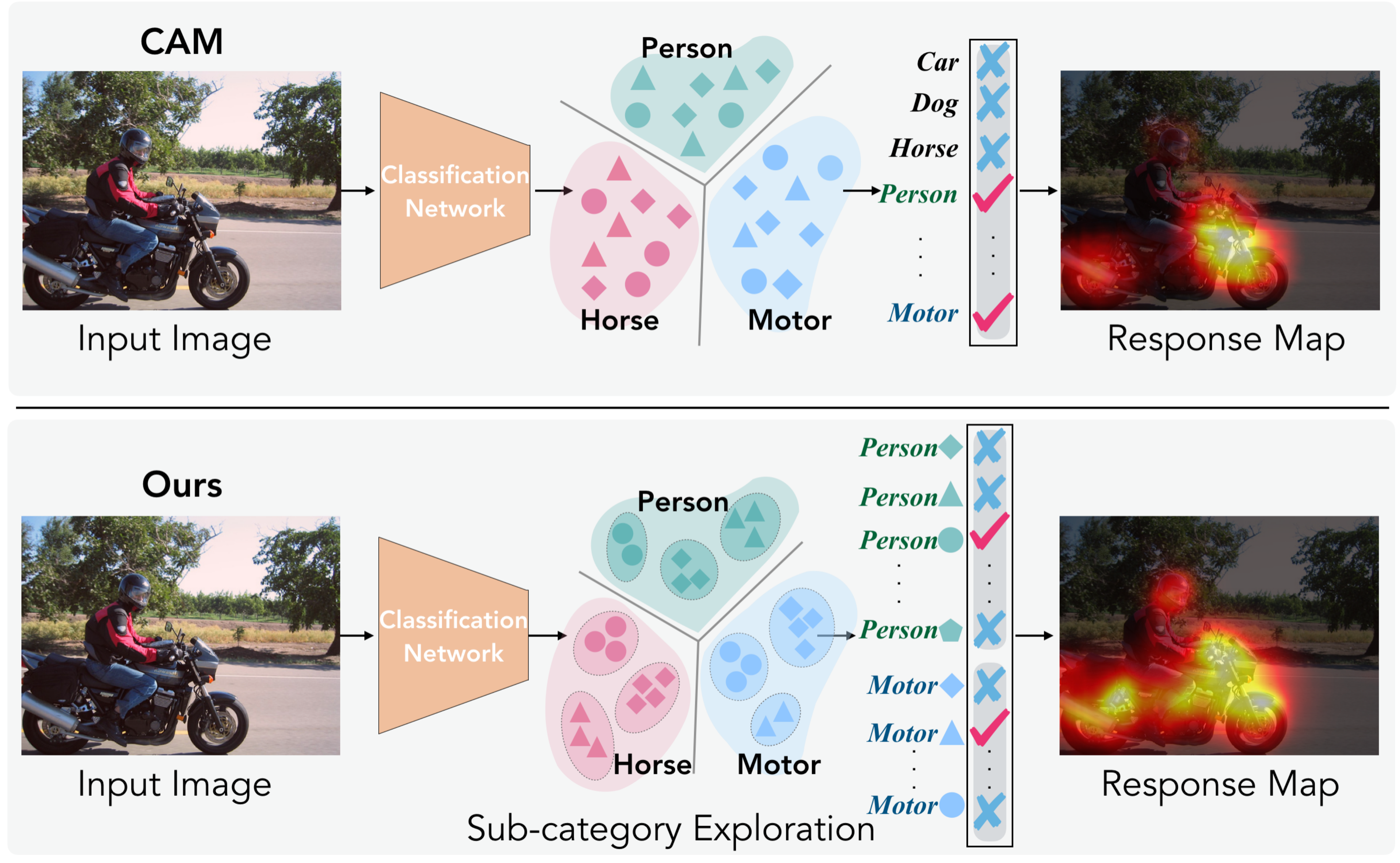}\\
	\caption{Existing weakly-supervised semantic segmentation methods based on image-level supervisions usually apply the class activation map (CAM) to obtain the response map as the initial prediction. 
	However, this response map can only highlight the discriminative parts of the object (top). 
	We propose a self-supervised task via sub-category exploration to enforce the classification network learn better response maps (bottom).
	}
	\label{fig:teaser}
 	\vspace{-3mm}
\end{figure}

In order to predict the initial response map for each category, numerous approaches based on the 
class activation map (CAM) model \cite{zhou2016learning}  
have been developed. 
Essentially, these methods train a classification network and use its learned weights in the classifier as the cues to compute weighted sums of feature maps, which can be treated as the response map.
However, such response maps may only focus on a portion of the object, instead of localizing the entire object (see top of Figure \ref{fig:teaser}).
%
%
One explanation is that the objective of the classifier does not need to ``see'' the entire object for optimizing the loss function.
This impairs the classifier's ability to locate the objects.

At the core of our technique is to impose a more challenging task to the network for learning better representations, while not jeopardizing the original objective.
To this end, we propose a simple yet effective method by introducing a self-supervised task that discovers sub-categories in an unsupervised manner, as illustrated at the bottom of Figure \ref{fig:teaser}.
Specifically, our task consists of two steps: 1) perform clustering on image features extracted from the classification network for each annotated parent class (e.g., 20 parent classes on the PASCAL VOC 2012 dataset \cite{PASCAL_VOC_2010_Data}), and 2) use the clustering assignment for each image as the pseudo label to optimize the sub-category objective.

On one hand, the parent classifier establishes a feature space through supervised training as the guidance for unsupervised sub-category clustering.
On the other hand, the sub-category objective provides additional gradients to enhance feature representations and leverage the sub-space of the original feature space to obtain better results.
As such, the classification model takes a more challenging task and is not limited to the easier objective of learning only the parent classifier.
Moreover, to ensure better convergence in practice, we iteratively alter the two steps of feature clustering and pseudo training the sub-category objective.

We conduct extensive experiments on the PASCAL VOC 2012 dataset \cite{PASCAL_VOC_2010_Data} to demonstrate the effectiveness of our method, with regard to generating better initial response maps to localize objects.
As a result, our approach leads to favorable performance for the final semantic segmentation results against state-of-the-art weakly-supervised approaches.
Furthermore, we provide extensive ablation studies and analysis to validate the robustness of our method.
Interestingly, we notice that the network is able to differentiate sub-categories with respect to their object size/type, context, and coexistence with other categories.
%
The main contributions of this work are summarized as follows:
\begin{itemize}
    \item We propose a simple yet effective method via a self-supervised task to enhance feature representations in the classification network. This  improves the initial class activation maps for weakly-supervised semantic segmentation as well.
    
    \item We explore the idea of sub-category discovery via iteratively performing unsupervised clustering and pseudo training on the sub-category objective in a self-supervised fashion.
    
    \item We present extensive study and analysis to show the efficacy of the proposed method, which significantly improves the quality of initial response maps and leads to better semantic segmentation results.

\end{itemize}

\begin{figure*}[t]
	\centering
	\includegraphics[width=0.95\linewidth]{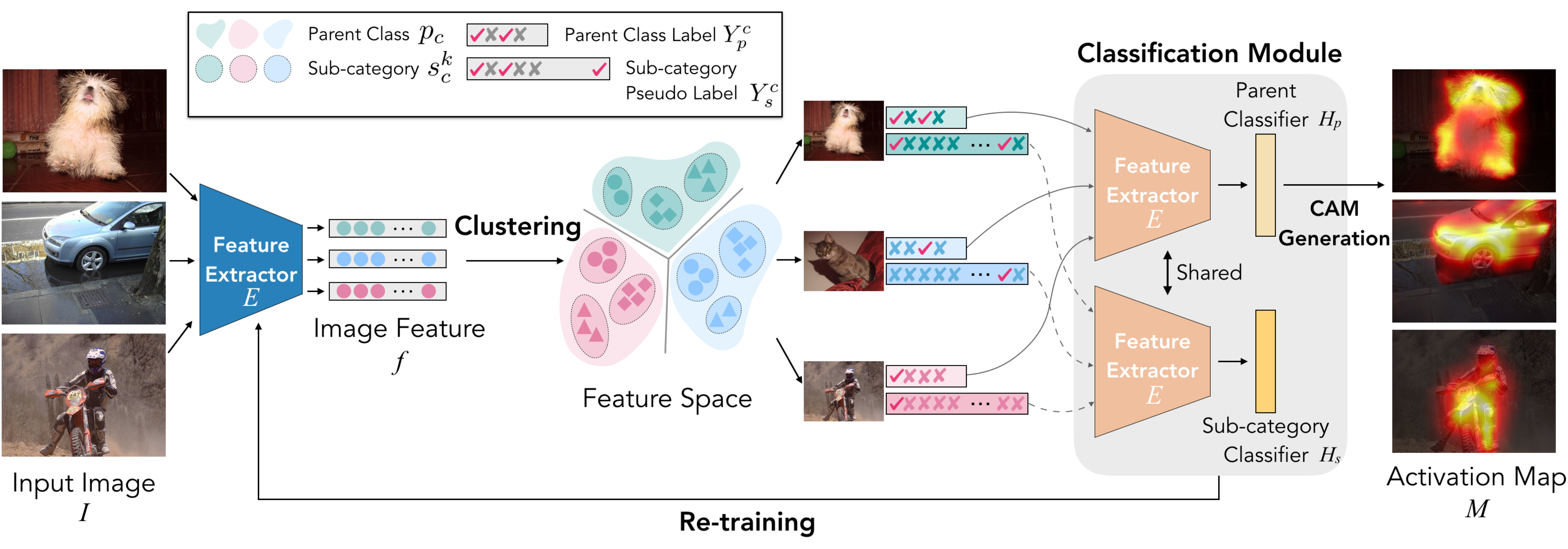}\\
	\caption{
		Proposed framework for generating the class activation map. 
		Given input images $I$, we first feed them into a feature extractor $E$ to obtain their features $f$.
		Then, we adopt unsupervised clustering on $f$ and obtain sub-category pseudo labels $Y_s$ for each image.
		Next, we train the classification network to jointly optimize the parent classifier $H_p$ with ground truth labels $Y_p$ for parent classes and the sub-category classifier $H_s$ using the sub-category pseudo labels obtained in the clustering stage.
		By iteratively performing unsupervised clustering on image features and pseudo training the classification module, we use the jointly optimized classification network to produce the final activation map $M$.
	}
	\label{fig: framework}
	\vspace{-3mm}
\end{figure*}

\section{Related Work}
Within the context of this work, we discuss methods for weakly-supervised semantic segmentation (WSSS) using image-level labels, including approaches that focus on initial prediction and refinement for generating pseudo ground truths. In addition, algorithms that are relevant to unsupervised representation learning are discussed in this section.

\paragraph{Initial Prediction for WSSS.}

Initial cues are essential for segmentation task since it can provide reliable priors to generate segmentation maps. The class activation map  \cite{zhou2016learning} is a widely used technique for localizing the object. It can highlight class-specific regions that are served as the initial cues. However, since the CAM model is trained by a classification task, it tends to activate to the small discriminative part of the object, leading to incomplete initial masks.

Several methods have been developed to alleviate this problem. Numerous approaches \cite{singh2017hide, wei2017object} deliberately hide or erase the region of an object, forcing models to seek more diverse parts. 
However, those methods either hide fixed-size patches randomly or require repetitive model training and response aggregation steps. 
A number of variants \cite{zhang2018adversarial, li2018tell} have been proposed to extend the initial response via an adversarial erasing strategy in an end-to-end training manner, yet such strategies may gradually expand their attention to non-object regions, leading to inaccurate attention maps. 
Recently, the SeeNet approach \cite{hou2018self} applies self-erasing strategies to encourage networks to use both object and background cues, which prevent the attention from including more background regions. 
Instead of using the erasing scheme, the FickleNet method \cite{lee2019ficklenet} introduces stochastic feature selection to obtain diverse combinations of locations on feature maps. By aggregating the localization maps, they acquire the initial cue that contains a larger region of the object. 

Different from the methods that mitigate the problem by discovering complementary regions via iterative erasing steps or consolidating attention maps, our proposed approach aims at enforcing the network to learn harder on a more challenging task via self-supervised sub-category exploration, thereby enhancing feature representations and improving the response map.
%

\paragraph{Response Refinement for WSSS.}
Numerous approaches \cite{ahn2018learning,fan2018cian, fan2018associating,huang2018weakly,kolesnikov2016seed,wang2018weakly,wei2018revisiting} are proposed to refine the initial cue via expanding the region of attention map.
The SEC method \cite {kolesnikov2016seed} proposes a loss function that constrains both global weighted rank pooling and low-level boundary to expand the localization map.
To improve the network training, the MCOF scheme \cite{wang2018weakly} uses a bottom-up and top-down framework which alternatively expands object regions and optimize the segmentation network, while the MDC method \cite{wei2018revisiting} expands the seeds by employing multiple branches of convolutional layers with different dilation rates.
Moreover, the DSRG approach \cite{huang2018weakly} refines initial localization maps by applying a seeded region growing method during the training of the segmentation network.
Other approaches are developed via affinity learning.
For instance, the AffinityNet \cite{ahn2018learning} considers pixel-wise affinity to propagate local responses to nearby areas, while \cite{fan2018cian, fan2018associating} explore cross-image relationships to obtain complementary information that can infer the predictions.

Nevertheless, initial seeds are still obtained from the CAM method. If these seeds only come from the discriminative parts of objects, it is difficult to expand regions into non-discriminative parts. Moreover, if the initial prediction produces wrong attention regions, applying the refinement step would cover even more inaccurate regions.
In this paper, we focus on improving the initial prediction, which leads to more accurate object localization and benefits the refinement step.

\paragraph{Unsupervised Representation Learning.}
Unsupervised learning has been widely studied in the computer vision community. One advantage is to learn better representations of images and apply learned features on any specific domain or dataset where annotations are not always available.
Self-supervised learning \cite{de1994learning} utilizes a pretext task to replace the labels annotated by humans with ``pseudo-labels'' directly computed from the raw input data.
A number of methods \cite{noroozi2016unsupervised, pathak2016context, paulin2015local} are developed but require expert knowledge to carefully design a pretext task that may lead to good transferable features.
To reduce the domain knowledge requirement, Coates and Ng \cite{coates2012learning} validate that feature-learning systems with K-means can be a scalable unsupervised learning module that can train a model of the unlabeled data for extracting meaningful features.
Furthermore, a recent approach \cite{caron2018deep} employs a clustering framework to extract useful visual features by 
alternating between clustering the image descriptors and updating the weights of the CNN by predicting the cluster assignments, in order to learn deep representations specific to domains where annotations are scarce.
In this work, we propose to learn a self-supervised method that explores the sub-category in the classification network, i.e., using unsupervised signal to enhance feature representations while improving initial response maps for weakly-supervised semantic segmentation.
%

\section{Weakly-supervised Semantic Segmentation}
In this section, we describe our framework for weakly-supervised semantic segmentation, including details of how we explore sub-categories to improve initial response maps and generate final semantic segmentation results.

\subsection{Algorithm Overview}

To obtain the initial response, we follow the common practice of training a classification network and utilize the CAM method \cite{zhou2016learning} to obtain our baseline model.
%
The CAM method typically only activates on discriminative object parts, which are not sufficient for the image classification task. To address this issue, We propose to integrate a more challenging task into the objective: self-supervised sub-category discovery, in order to enforce the network to learn from more object parts.
%

Firstly, for each annotated parent class, we determine $K$ sub-categories by applying K-means clustering on image features.
With the clustering results, we then assign each image with a pseudo label, which is identified as the index of the sub-category. Finally, we construct a sub-category objective to jointly train the classification network.
By iteratively updating the feature extractor, two classifiers, and sub-category pseudo labels, the enhanced features representations lead to better classification, and thereby gradually produce response maps that attain to more complete regions of the objects. 
The overall process is illustrated in Figure \ref{fig: framework}.
Then, we use the method in \cite{ahn2018learning} to expand response maps, which are used as pseudo ground truths to train the segmentation network.
Also note that, our method focuses on the initial prediction, so it is not limited to certain region expansion or segmentation training methods.

\paragraph{Preliminaries: Initial Response via CAM.}
We adopt the CAM to generate the initial response using a typical classification network, whose architecture consists of convolutional layers as the feature extractor $E$, followed by global average pooling (GAP) and one fully-connected layer $H_p$ as the output classifier.
Given an input image $I$, the network is trained with image-level labels $Y_p$ using a multi-label classification loss $\mathcal{L}_p$, following \cite{zhou2016learning}.
After training, the activation map $M$ for each category $c$ can be obtained via directly applying classifier $H_p$ on the feature maps $f = E(I)$:
\begin{equation}
    M^c(x,y) = \theta_p^{c \top} f(x,y),
    \label{eq:cam}
\end{equation}
where $\theta_p^c$ is the classifier weight for the category c, and $f(x,y)$ is the feature at pixel $(x,y)$. The response map is further normalized by the maximum value in $M^c$.
%


\subsection{Sub-category Exploration}
The activation map for each image using \eqref{eq:cam} provides typically highlights only the discriminative object parts.
However, from the perspective of a classifier, discovering the most discriminative part of the object is already sufficient for optimizing the loss function $\mathcal{L}_p$ in classification.
As the learning objective is based on the classification scores, it is inevitable for the CAM model to generate 
incomplete attention maps.
To address this issue, we integrate a self-supervised scheme to enhance feature representations $f$ while improving the response maps via exploring the sub-category information, in which $f$ appears to be an important cue to compute the activation map via \eqref{eq:cam}.

%

%
\paragraph{Sub-Category Objective.}
To assign a more challenging problem to the classification model, we introduce a task to discover sub-categories in an unsupervised manner.
For each parent class $p_c$, we define $K$ sub-categories $s_c^k$, where $k = \{1,2, ... ,K\}$.
%
%
For each image $I$ with the parent label $Y_p^c$ in $\{0, 1\}^{c}$, the corresponding sub-category label for the category $c$ is denoted as $Y_s^{c,k}$ in $\{0, 1\}^{k}$.
We also note that, if the label of one parent class does not exist (i.e., $Y_p^c = 0$), the labels of all sub-categories would be also 0, i.e., $Y_s^{c,k} = 0, k = \{1,2,...,K\}$.
Our objective is to learn a sub-category classifier $H_s$ parameterized with $\theta_s$, while sharing the same feature extractor $E$ with $H_p$. 
Similar to the parent classification loss $\mathcal{L}_p$, we adopt the standard multi-label classification loss $\mathcal{L}_s$ with a larger and fine-grained label space $Y_s$.

\vspace{-3mm}
\paragraph{Sub-category Discovery.}
As there is no ground truth label for sub-category to directly optimize the above sub-category objective $\mathcal{L}_s$, we generate pseudo labels via unsupervised clustering. 
Specifically, we perform clustering for each parent class on image features extracted from the feature extractor $E$.
The clustering objective for each class $c$ can be written as:
\begin{equation}
    \min_{D \in \mathbb{R}^{d \times k}} \frac{1}{N^c} \sum_{i=1}^{N^c} \min_{Y_s^c} ||f - TY_s^c||_2^2, \;\;\text{s.t.}, Y_{s}^{c \top} 1_k = 1,
    \label{eq:clustering}
\end{equation}
where $T$ is a $D \times K$ centroid matrix, $N^c$ is the number of images containing the class $c$, and $f = E(I) \in \mathbb{R}^D$ is the extracted feature.
We use the clustering assignment $Y_s^c$ for each image as the sub-category pseudo label to optimize $\mathcal{L}_s$.
%

\vspace{-3mm}
\paragraph{Joint Training.}
After obtaining sub-category pseudo labels $Y_s$ from the above clustering process, we jointly optimize the feature representations $f = E(I)$ and two classifiers, i.e., $H_p$ and $H_s$:
\begin{equation}
    \min_{\theta_p, \theta_s} \frac{1}{N} \sum_{i=1}^{N} \mathcal{L}_p(H_p(f_i), Y_p) + \lambda \mathcal{L}_s(H_s(f_i), Y_s),
    \label{eq:joint}
\end{equation}
%
%
where $N$ is the total number of images and $\lambda$ is weight to balance two loss functions.
With this method, the parent classification learns a feature space through supervised training via $\mathcal{L}_p$, while the sub-category objective $\mathcal{L}_s$ explores the feature sub-space and provides additional gradients to enhance feature representations $f$, which is used to compute CAM via \eqref{eq:cam}.

\begin{algorithm}[!t]
\vspace{0.1in}
\caption{Learning Sub-category Discovery for CAM}\label{alg:wwws}
\begin{algorithmic}
\State \textbf{Input:} Image $I$; Parent Label $Y_p$; Category Number $C$; \\ Sub-category Number $K$
\State \textbf{Output:} Class Activation Map $M^c$\\

\State \textbf{Model:} Feature extractor $E$; Parent Classifier $(H_p; \theta_p)$; \\ Sub-category Classifier $(H_s; \theta_s)$ \\

\State Optimize $\{E, H_p\}$ with $Y_p$ via $\mathcal{L}_p$
\While{Training}

\State Extract features via $f = E(I)$

\For{$c \gets 1$ to $C$}
\State Generate pseudo labels $Y_s^c$ with $f$ via \eqref{eq:clustering}
\EndFor

\State Optimize $\{E, H_p, H_s\}$ with $\{Y_p, Y_s\}$ via \eqref{eq:joint}

\EndWhile
\\
Compute $M^c$ via \eqref{eq:cam}
\end{algorithmic}
\end{algorithm}
\vspace{-1mm}

\paragraph{Iterative Optimization.}
%
%
The proposed unsupervised clustering scheme in \eqref{eq:clustering} relies on the feature $f$ to discover sub-category pseudo labels. 
As such, the learned features via only the objective $\mathcal{L}_p$ could be less discriminative for the clustering purpose.
To mitigate this issue, we adopt an iterative training method by alternatively updating \eqref{eq:clustering} and \eqref{eq:joint}.
Therefore, features $f$ are first enhanced through the sub-category objective, and in turn facilitate the clustering process to generate better pseudo ground truths, which are then used to learn better feature representations in network training.
The overall optimization for generating final class activation maps is summarized in Algorithm \ref{alg:wwws}.

%



\subsection{Implementation Details}
In this section, we describe implementation details of the proposed framework and the following procedures to produce final semantic segmentation results. 
All the source code and trained models are available at \url{https://github.com/Juliachang/SC-CAM}.
%

\paragraph{Classification Network.}
In this work, the ResNet-38 architecture \cite{wu2019wider} is used for the CAM model, and the training procedure is similar to that in \cite{ahn2018learning}.  
%
The network consists of 38 convolution layers with wide channels, followed by a $3 \times 3$ convolution layer with 512 channels for better adaptation to the classification task, a global average pooling layer for feature aggregation, and two fully-connected layers for image and sub-category classification, respectively.
The model is pre-trained on the ImageNet \cite{deng2009imagenet} and is then finetuned on the PASCAL VOC 2012 dataset.
%
%
We use the typical techniques based on the horizontal flip, random cropping, and color jittering operations to augment the training data set. 
We also randomly scale input images to impose scale invariance in the network.

We implement the proposed framework with PyTorch and train on a single Titan X GPU with 12 GB memory. 
To train the classification network, we use the Adam optimizer \cite{kingma2014adam} with initial learning rate of 1e-3 and the weight decay of 5e-4.
In practice, we use $\lambda = 5$ and $K=10$ in all the experiments unless specified otherwise.
For iterative training, we empirically find that the model converges after training for 3 rounds.
In the experimental section, we show studies for the choice of $K$ and iterative training results.
%

\begin{table}[!t]
	\caption{Performance comparison in mIoU (\%) for evaluating activation maps on the PASCAL VOC training and validation sets.}
	\vspace{-1mm}
	\small
	\centering
	\renewcommand{\arraystretch}{1.1}
	\setlength{\tabcolsep}{6pt}
	\begin{tabular}{lcc|cc}
		    \toprule
	        & \multicolumn{2}{c}{Training Set} & \multicolumn{2}{c}{Validation Set} \\
		    \midrule
		    Method & CAM & CAM+RW & CAM & CAM+RW \\
		    \midrule
		    AffinityNet \cite{ahn2018learning} & 48.0 & 58.1 & 46.8 & 57.0 \\
		    Ours & 50.9 & 63.4 & 49.6 & 61.2 \\
		\bottomrule
	\label{table: compare_cam}
	\end{tabular}
	\vspace{-8mm}
\end{table}

\paragraph{Semantic Segmentation Generation.}
Based on the response map generated by our method as in Algorithm \ref{alg:wwws}, we adopt the random walk method via affinity \cite{ahn2018learning} to refine the map as pixel-wise pseudo ground truths for semantic segmentation.
In addition, as a common practice, we use dense conditional random fields (CRF) \cite{crf} to further refine the response to obtain better object boundaries.
To train the segmentation network, we utilize the Deeplab-v2 framework \cite{deeplab} with the ResNet-101 architecture \cite{He_2016_CVPR} as the backbone model.
%

\begin{figure*}[t]
	\centering
	\includegraphics[width=1\linewidth]{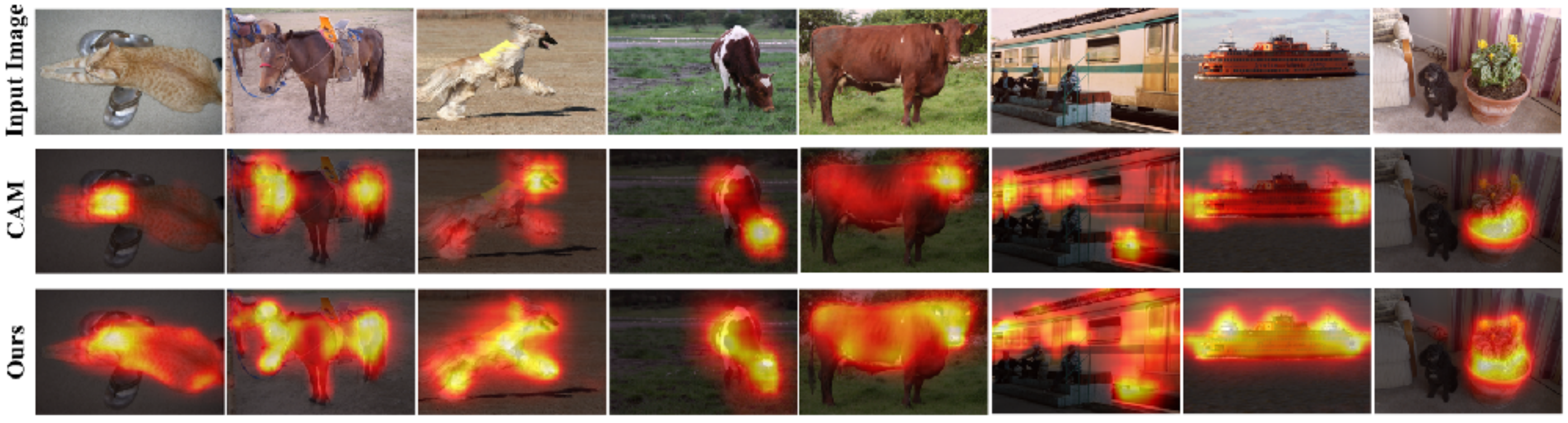}\\
	\caption{Sample results of initial responses. 
	Our method often generates the response map that covers larger region of the object (i.e., attention on the body of the animal), while the response map produced by CAM \cite{zhou2016learning} tends to highlight small discriminative parts.}
	\label{fig: cam_viz}
	\vspace{-1mm}
\end{figure*}

\begin{figure}[!t]
	\centering
	\includegraphics[width=0.85\linewidth]{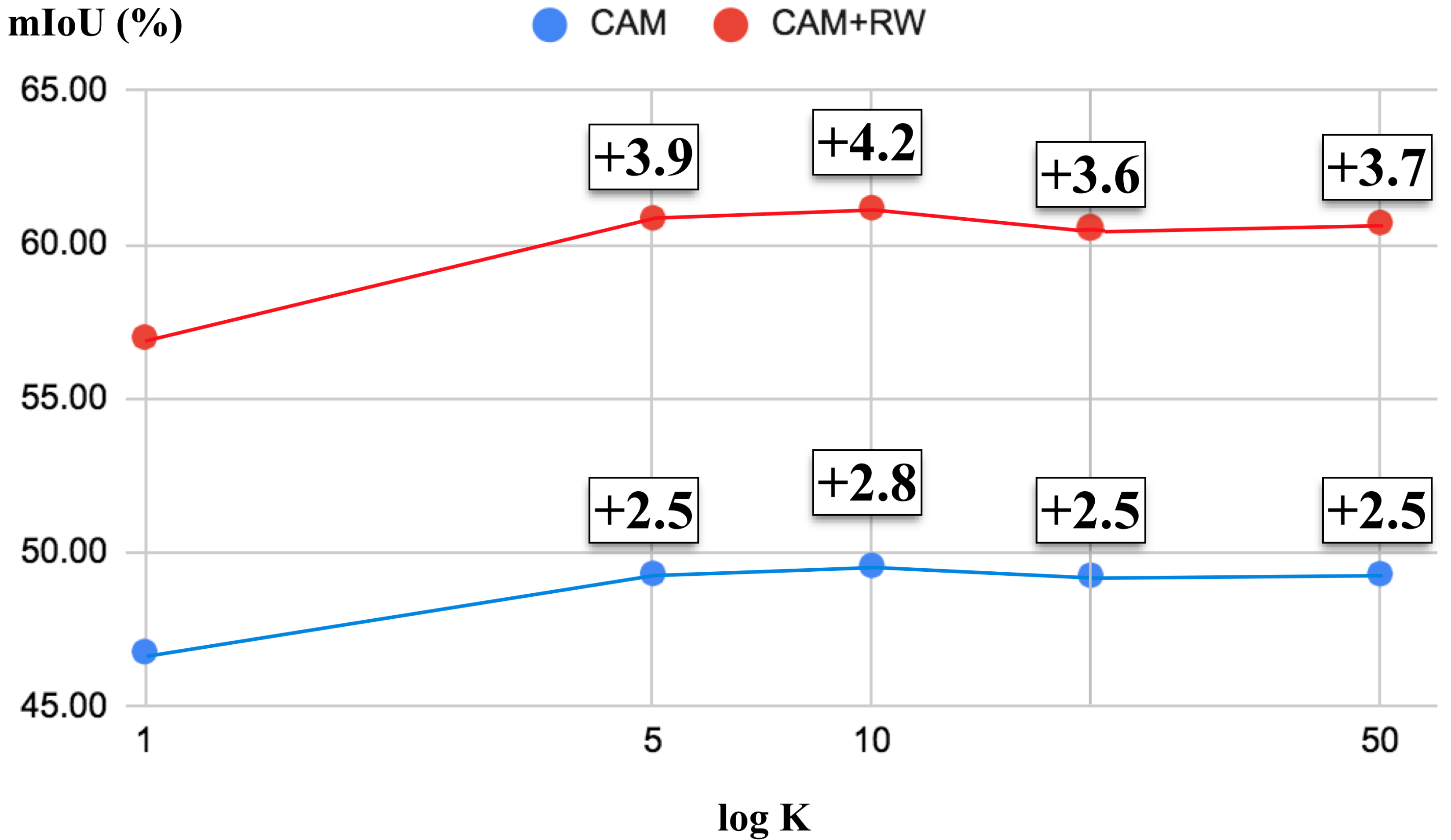}\\
	\vspace{-1mm}
	\caption{Ablation study for $K$. We show that the proposed method performs robustly with respect to $K$ and is consistently better than the original CAM that did not apply clustering to discover sub-categories. We mark the value of mIoU of the original CAM at $K = 1$ and the improved mIoUs are presented.}
	\label{fig: k}
	\vspace{-3mm}
\end{figure}

\section{Experimental Results}
In this section, we first present the main results and analysis of the initial response generated by our method.
Second, we show the final semantic segmentation performance on the PASCAL VOC dataset \cite{PASCAL_VOC_2010_Data} against the state-of-the-art approaches.
%
More results can be found in the supplementary material. 

%
\subsection{Evaluated Dataset and Metric}
We evaluate the proposed approach on the PASCAL VOC 2012 semantic segmentation benchmark \cite{PASCAL_VOC_2010_Data} which contains 21 categories, including one background class. 
Each image contains one or multiple object classes. Following previous weakly-supervised semantic segmentation methods, we use augmented 10,528 training images present in \cite{hariharan2011semantic} along with their image-level labels to train the network. 
To evaluate the training set, we use the set without augmentation which has 1,464 examples. 
We adopt 1,449 images in the validation set and 1,456 images in the test set to compare our results with other methods. 
For all experiments, the mean Intersection-over-Union (mIoU) ratio is used as the evaluation metric. 
The results for the test set are obtained from the official PASCAL VOC evaluation website.



\subsection{Improvement on Initial Response}
In Table \ref{table: compare_cam}, we show the mean IoU of the segments computed using the CAM on both the training and validation sets.
%
We present results after applying the refinement step to the activation map, i.e., CAM + random walk (CAM + RW).
Table \ref{table: compare_cam} shows that our approach significantly improves the IoU over AffinityNet \cite{ahn2018learning} by almost 3$\%$ using CAM and more than 4$\%$ for CAM+RW.
The improved initial response maps facilitate the downstream task in generating pixel-wise pseudo ground truths for training the semantic segmentation model.

In Figure \ref{fig: cam_viz}, we show comparisons of generated CAMs by the conventional classification loss $\mathcal{L}_p$ \cite{zhou2016learning} and the proposed method via sub-category discovery summarized in Algorithm \ref{alg:wwws}.
Visual results show that our method is able to localize more complete object regions, while the original CAM only focuses on discriminative object parts.
We also note that this is essentially critical for the refinement stage that takes the response map as the input.



\begin{table}[!t]
	\caption{Segmentation quality of the initial response at different rounds of training on the PASCAL VOC 2012 validation set. We show there is a gradual improvement on both mIoU and F-Score metrics.}
	\vspace{-2mm}
	\small
	\centering
	\renewcommand{\arraystretch}{1.1}
	\setlength{\tabcolsep}{6pt}
	\begin{tabular}{lcc}
		\toprule
		Round & mIoU (\%) $\uparrow$ & F-Score $\uparrow$ \\
		\midrule
	
		\#0 (CAM) & 46.8 & 65.1 \\
		\#1 & 48.0 & 65.6 \\
		\#2 & 48.7 & 66.6 \\
		\#3 & \textbf{49.6} & \textbf{67.0} \\
		
		\bottomrule
	\label{table: iter_improve}
	\end{tabular}
	\vspace{-5mm}
\end{table}

\subsection{Ablation Study and Analysis}
To demonstrate how our method helps improve feature representations and allow the network pay more attention to other object parts via exploiting the sub-category information, we present extensive analysis in this section.
Here, all the experimental results are based on the PASCAL VOC validation set.

\vspace{-2mm}
\paragraph{Effect of Sub-Category Number $K$.}
We first study how the sub-category number $K$  affect the performance of the proposed method. 
%
In Figure \ref{fig: k}, we use $K = \{5, 10, 20, 50\}$, and show that the proposed method performs robustly with respect to $K$ (within a wide range) and consistently better than the original CAM method (i.e., $K = 1$).
The results also validate the necessity and importance of using more sub-categories (i.e., $K>1$) to generating better response maps. 
%
%
Considering the efficiency and accuracy, we use $K = 10$ for each parent class in all the experiments.
\textcolor{black}{As a future work, it is of great interest to develop an adaptive method to determine the sub-category number \cite{sarfraz2019efficient}, which can reduce the redundant sub-categories and make the approach more efficient.}

%

\vspace{-2mm}
\paragraph{Iterative Improvement.}
To demonstrate the effectiveness of our iterative training process, we show the gradual improvement on the segment quality in Table \ref{table: iter_improve}.
We present the results of mIoU and F-Score that accounts for both the recall and precision measurements, in which they are important cues to validate whether the activation map is able to cover object parts.
Compared to the results in round \#0, which is the original CAM, our method gradually improves both metrics as training more rounds.

\begin{figure*}[t]
	\centering
	\includegraphics[width=1\linewidth]{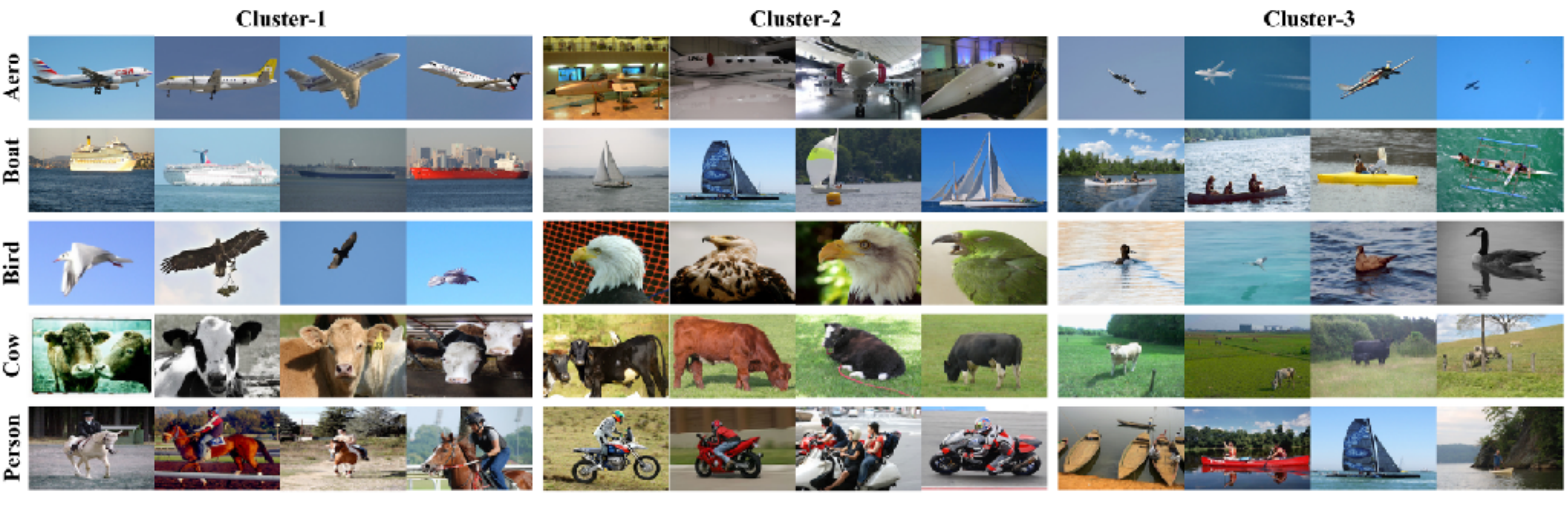}\\
	\vspace{-2mm}
	\caption{Clustering results of the last round model (\#3). We show 3 clusters for each parent class and demonstrate that our learned features are able to cluster objects based on their size (\textit{Aeroplane, Bird, Cow}), context (\textit{Aeroplane, Bird, Person}), type (\textit{Boat, Bird}), pose (\textit{Cow}), and interaction with other categories (\textit{Person}).
	}
	\label{fig: cluster_viz}
\end{figure*}

\begin{figure*}[t]
	\centering
	\includegraphics[width=1\linewidth]{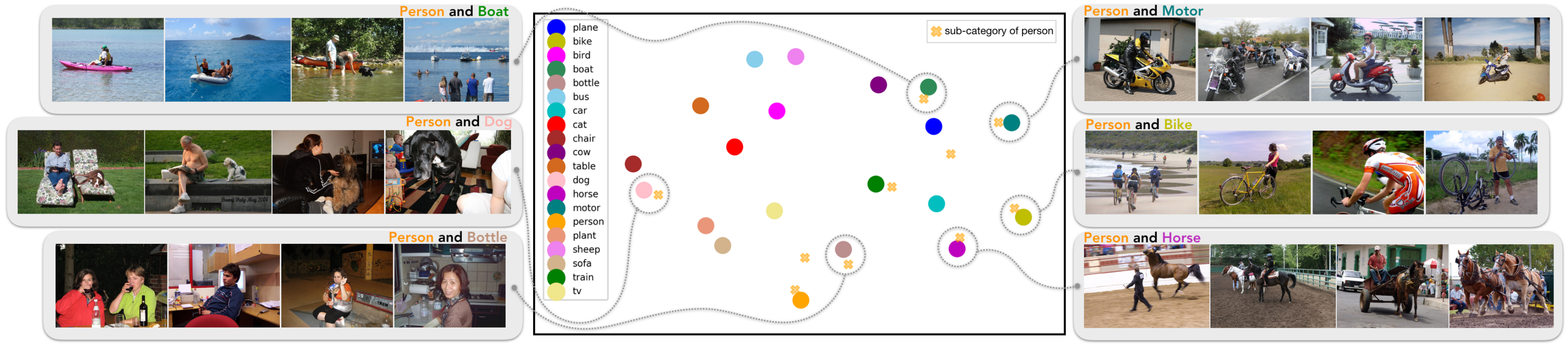}\\
	\caption{Visualizations of weights based on the 
	t-SNE method that illustrates the relationships on semantic-level between parent classifier and the person sub-category classifier. We show that one person sub-category is usually close to one parent class, as they often co-appear in the same image, as shown in example images on two sides.}
	\label{fig: weight_viz}
\end{figure*}

\paragraph{Clustering Results.}
Since the ground truth labels are not available for sub-categories, we present visualizations of clustering results in Figure \ref{fig: cluster_viz} to measure the quality, in which each parent class shows 3 example clusters. 
Our method is able to cluster objects based on their size (\textit{Aeroplane, Bird, Cow}), context (\textit{Aeroplane, Bird, Person}), type (\textit{Boat, Bird}), pose (\textit{Cow}), and interaction with other categories (\textit{Person}).
For instance, persons with different categories, e.g., horse, motobike, and boat, are clustered into different groups.
This visually validates that our learned feature representations are enhanced via the sub-category objective in an unsupervised manner.
More visual comparisons are presented in the supplementary material.


\paragraph{Weight Visualization.}
%
In order to understand how our learning mechanism improves the clustering quality, we visualize the distribution of the classifier weights, i.e., $\theta_p$ and $\theta_s$, via t-SNE \cite{maaten2008visualizing}.
As such, we are able to find the relationship between the parent classifier $H_p$ and the sub-category module $H_s$.
Figure \ref{fig: weight_viz} shows the visualization of weights, in which we take the sub-categories of person (denoted as yellow cross symbols) as the example, since the person category has more interactions with other parent classes (denoted as solid circles).
It illustrates that one person sub-category is often close to one parent class, e.g., sub-category \textit{person} and parent class \textit{bike}, which makes sense as those two categories usually co-appear in the same image (see example images in Figure \ref{fig: weight_viz} on two sides).
%



\begin{figure*}[!t]
	\centering
	\includegraphics[width=1\linewidth]{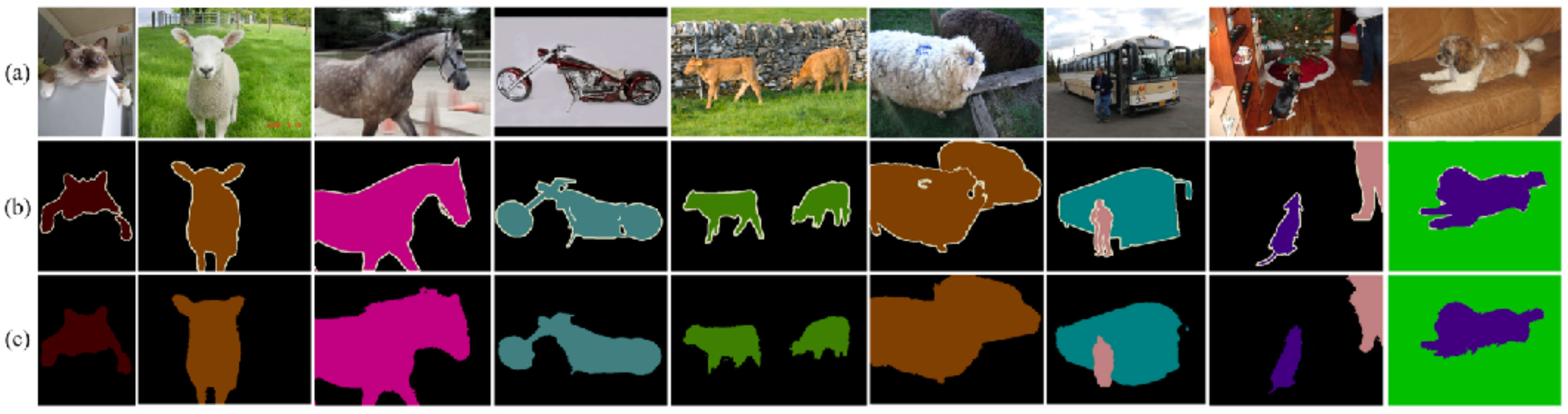} \\
	\vspace{-2mm}
	\caption{Qualitative results on the PASCAL VOC 2012 validation set.
	(a) Input images. (b) Ground truth. (c) Our results.
	}
	\label{fig: segmap_viz}
	
\end{figure*}

\vspace{2mm}
\begin{table*} [t]
	\caption{Semantic segmentation performance on the PASCAL VOC 2012 validation set. Bottom group contains results with CRF refinement, while the top group is without CRF. Note that 11/20 classes obtain improvements using our approach w/ CRF. The best three results are in \textcolor{red}{red}, \textcolor{PineGreen}{green} and \textcolor{blue}{blue}, respectively.}
	
	\vspace{-1mm}
	\footnotesize
	\centering
	\renewcommand{\arraystretch}{1.3}
	\setlength{\tabcolsep}{2pt}
	\begin{tabular}{lccccccccccccccccccccc|c}
		\toprule
		
		
		Method & bkg & aero & bike & bird & boat & bottle & bus & car & cat & chair & cow & table & dog & horse & motor & person & plant & sheep & sofa & train & tv & mIoU \\
		\midrule

		AffinityNet  \cite{ahn2018learning} & 88.2 & 68.2 & \textcolor{blue}{30.6} & \textcolor{PineGreen}{81.1} & 49.6 & 61.0 & 77.8 & 66.1 & 75.1 & \textcolor{blue}{29.0} & 66.0 & 40.2 & \textcolor{PineGreen}{80.4} & 62.0 & \textcolor{PineGreen}{70.4} & \textcolor{red}{73.7} & \textcolor{blue}{42.5} & 70.7 & \textcolor{PineGreen}{42.6} & \textcolor{blue}{68.1} & 51.6 & 61.7 \\
		
		Ours (w/o CRF) & 88.1 & 49.6 & 30.0 & 79.8 & \textcolor{blue}{51.9} & \textcolor{PineGreen}{74.6} & \textcolor{PineGreen}{87.7} & 73.7 & \textcolor{PineGreen}{85.1} & \textcolor{PineGreen}{31.0} & \textcolor{blue}{77.6} & \textcolor{PineGreen}{53.2} & \textcolor{blue}{80.3} & \textcolor{PineGreen}{76.3} & \textcolor{blue}{69.6} & 69.7 & 40.7 & 75.7 & \textcolor{PineGreen}{42.6} & 66.1 & \textcolor{blue}{58.2} & \textcolor{blue}{64.8} \\
		
		\midrule
		
		MCOF \cite{wang2018weakly}  & 87.0 &  \textcolor{red}{78.4} & 29.4 & 68.0 & 44.0 & 67.3 & 80.3 & \textcolor{blue}{74.1} & 82.2 & 21.1 & 70.7 & 28.2 & 73.2 & 71.5 & 67.2 & 53.0 & \textcolor{PineGreen}{47.7} & 74.5 & 32.4 & \textcolor{PineGreen}{71.0} & 45.8 & 60.3 \\
		
		Zeng et al. \cite{zeng2019joint} &  \textcolor{red}{90.0} & \textcolor{PineGreen}{77.4} &  \textcolor{red}{37.5} & \textcolor{blue}{80.7} &  \textcolor{red}{61.6} & 67.9 & 81.8 & 69.0 & \textcolor{blue}{83.7} & 13.6 & \textcolor{PineGreen}{79.4} & 23.3 & 78.0 & \textcolor{blue}{75.3} &  \textcolor{red}{71.4} & 68.1 & 35.2 & \textcolor{blue}{78.2} & 32.5 & \textcolor{red}{75.5} & 48.0 & 63.3\\ 
		
		FickleNet \cite{lee2019ficklenet} & 
		\textcolor{PineGreen}{89.5} & \textcolor{blue}{76.6} & \textcolor{PineGreen}{32.6} & 74.6 & 51.5 & \textcolor{blue}{71.1} & \textcolor{blue}{83.4} & \textcolor{PineGreen}{74.4} & 83.6 & 24.1 & 73.4 & \textcolor{blue}{47.4} & 78.2 & 74.0 & 68.8 &  \textcolor{PineGreen}{73.2} &  \textcolor{red}{47.8} &  \textcolor{red}{79.9} & \textcolor{blue}{37.0} & 57.3 & \ \textcolor{red}{64.6} & \textcolor{PineGreen}{64.9} \\

		Ours (w/ CRF) & \textcolor{blue}{88.8} & 51.6 & 30.3 & \textcolor{red}{82.9} & \textcolor{PineGreen}{53.0} & \textcolor{red}{75.8} &  \textcolor{red}{88.6} &  \textcolor{red}{74.8} &  \textcolor{red}{86.6} &  \textcolor{red}{32.4} &  \textcolor{red}{79.9} &  \textcolor{red}{53.8} &  \textcolor{red}{82.3} & 
		\textcolor{red}{78.5} & \textcolor{PineGreen}{70.4} & \textcolor{blue}{71.2} & 40.2 & \textcolor{PineGreen}{78.3} & \textcolor{red}{42.9} & 66.8 & \textcolor{PineGreen}{58.8} &  \textcolor{red}{66.1} \\

		\bottomrule
	\label{table: compare_sota_detail}
	\end{tabular}
	\vspace{-5mm}
\end{table*}

\begin{table}[!t]
\caption{Comparison of weakly-supervised semantic segmentation methods on the PASCAL VOC 2012 val and test sets. In addition, we present methods that aim to improve the initial response with $\checkmark$ in the ``Init. Res.'' column.}
\label{tab:orbital_data}
\centering
\small
\renewcommand{\arraystretch}{1.1}
\setlength{\tabcolsep}{3.5pt}
\begin{tabular}{lcccc} 
\toprule
\text{Method} & {Backbone} & {Init. Res.} & {Val} & {Test} \\
\midrule
    MCOF \textsubscript{CVPR'18} \cite{wang2018weakly} & ResNet-101 &  & 60.3 & 61.2 \\

    DCSP \textsubscript{BMVC'17} \cite{chaudhry2017discovering} & ResNet-101 &  & 60.8 & 61.9 \\

    DSRG \textsubscript{CVPR'18} \cite{huang2018weakly} & ResNet-101 &  & 61.4 & 63.2 \\

    AffinityNet \textsubscript{CVPR'18} \cite{ahn2018learning} & Wide ResNet-38 & & 61.7 & 63.7 \\

    SeeNet \textsubscript{NIPS'18} \cite{hou2018self} & ResNet-101 & \checkmark & 63.1 & 62.8 \\

    Zeng \textit{et al} \textsubscript{ICCV'19} \cite{zeng2019joint} &  DenseNet-169 &  & 63.3 & 64.3 \\

    BDSSW \textsubscript{ECCV'18} \cite{fan2018associating} & ResNet-101 & & 63.6 & 64.5 \\

    
    OAA \textsubscript{ICCV'19} \cite{jiang2019integral} & ResNet-101 & \checkmark & 63.9 & 65.6 \\

    CIAN \textsubscript{CVPR'19} \cite{fan2018cian} & ResNet-101 &  & 64.1 & 64.7 \\

    FickleNet \textsubscript{CVPR'19} \cite{lee2019ficklenet} & ResNet-101 & \checkmark & 64.9 & 65.3 \\
    
    
    Ours & ResNet101 & \checkmark & \textbf{66.1} & \textbf{65.9} \\
    
\bottomrule
\label{table: compare_sota}
\end{tabular}
\vspace{-8mm}
\end{table}

 \vspace{-2mm}
\subsection{Semantic Segmentation Performance}
%
%
After generating the pseudo ground truths as the results in Table \ref{table: compare_cam} (i.e., CAM + RW), we use them to train the semantic segmentation network.
We first compare our method with recent work using the ResNet-101 backbone or other similarly powerful ones in Table \ref{table: compare_sota}.
On both validation and testing sets, the proposed algorithm performs favorably against the  state-of-the-art approaches.
We also note that, most methods focus on improving the refinement stage or network training, while ours improves the initial step to generate better object response maps.

In Table \ref{table: compare_sota_detail}, we show detailed results for each category on the validation set.
We compare two groups of results with (bottom) or without (top) applying the CRF \cite{crf} refinement to the final segmentation outputs.
Compared to the recent FickleNet \cite{lee2019ficklenet} method that also focuses on improving the initial response map, the proposed algorithm performs favorably for the segmentation task in terms of the mean IoU.
We also note that, our results without applying CRF (mIoU as $64.8\%$) already achieves similar performance compared with the FickleNet (mIoU as $64.9\%$).
%
%
In Figure \ref{fig: segmap_viz}, we present some examples of the final semantic segmentation results, and show that our results are close to the ground truth segmentation.



\section{Conclusions}
In this paper, we propose a simple yet effective approach to improve the class activation maps by introducing a self-supervised task to discover sub-categories in an unsupervised manner. Without bells and whistles, our approach performs favorably against existing weakly-supervised semantic segmentation methods.
Specifically, we develop an iterative learning scheme by running clustering on image features for each parent class and train the classification network on sub-category objectives.
Unlike other existing schemes that aggregate multiple response maps, our approach generates better initial predictions without introducing extra complexity or inference time to the model.
We conduct extensive experimental analysis to demonstrate the effectiveness of our approach via exploiting the sub-category information.
Finally, we show that our algorithm produces better activation maps, thereby improving the final semantic segmentation performance.

\vspace{-3mm}
{\flushleft \bf{Acknowledgments.}}
This work is supported in part by the NSF CAREER Grant \#1149783, and gifts from eBay and Google.

{\small
\bibliographystyle{ieee_fullname}
\bibliography{mybib}
}

\end{document}